\pdfoutput=1

\documentclass[runningheads]{llncs}

\usepackage{eccv}

\usepackage{eccvabbrv}

\usepackage{graphicx}
\usepackage{booktabs}

\usepackage[accsupp]{axessibility}  

\usepackage{hyperref}

\usepackage{orcidlink}

\begin{document}

\title{HSEmotion Team at the 7th ABAW Challenge: Multi-Task Learning and Compound Facial Expression Recognition} 

\titlerunning{HSEmotion Team at the 7th ABAW Challenge}

\author{Andrey V. Savchenko\inst{1,2}\orcidlink{0000-0001-6196-0564}}

\authorrunning{A.V.~Savchenko.}

\institute{Sber AI Lab, Moscow, Russia \and
HSE University, Laboratory of Algorithms and Technologies for Network Analysis, Nizhny Novgorod, Russia\\
\email{avsavchenko@hse.ru}}

\maketitle

\begin{abstract}
In this paper, we describe the results of the HSEmotion team in two tasks of the seventh Affective Behavior Analysis in-the-wild (ABAW) competition, namely, multi-task learning for simultaneous prediction of facial expression, valence, arousal, and detection of action units, and compound expression recognition. We propose an efficient pipeline based on frame-level facial feature extractors pre-trained in multi-task settings to estimate valence-arousal and basic facial expressions given a facial photo. We ensure the privacy-awareness of our techniques by using the lightweight architectures of neural networks, such as MT-EmotiDDAMFN, MT-EmotiEffNet, and MT-EmotiMobileFaceNet, that can run even on a mobile device without the need to send facial video to a remote server. It was demonstrated that a significant step in improving the overall accuracy is the smoothing of neural network output scores using Gaussian or box filters. It was experimentally demonstrated that such a simple post-processing of predictions from simple blending of two top visual models improves the F1-score of facial expression recognition up to 7\%. At the same time, the mean Concordance Correlation Coefficient (CCC) of valence and arousal is increased by up to 1.25 times compared to each model's frame-level predictions. As a result, our final performance score on the validation set from the multi-task learning challenge is 4.5 times higher than the baseline (1.494 vs 0.32).

  \keywords{Affective Behavior Analysis in-the-Wild (ABAW) \and facial expression recognition \and valence-arousal prediction \and action unit detection \and compound expression recognition \and lightweight neural networks \and smoothing of frame-wise predictions}
\end{abstract}

\section{Introduction}
\label{sec:intro}

In recent years, the development of robust and accurate models for facial expression recognition has garnered significant attention due to its applications in various fields, such as human-computer interaction, mental health assessment, and surveillance. However, building models that are accurate in cross-dataset settings, fair, explainable, and robust remains a challenging task. Traditional approaches often struggle with biases introduced by unevenly distributed training data, resulting in models that perform well on average but poorly on underrepresented subgroups. Moreover, the necessity of balancing high performance with ethical considerations like privacy and explainability further complicates the development of these models. Ensuring that the models perform well across all demographic subgroups, and in unconstrained environments (i.e., in-the-wild conditions), poses additional difficulties. These challenges are highlighted by the organizers of a sequence of Affective Behavior Analysis in-the-wild (ABAW) contests~\cite{kollias2020analysing,kollias2021affect,kollias2021analysing,kollias2023abaw2} based on AffWild~\cite{zafeiriou2017aff,kollias2019deep} and AffWild2~\cite{kollias2019expression} datasets.

The 7th ABAW competition~\cite{kollias20247th} highlights these challenges by tasking participants with two tasks: 1) multi-task learning (MTL)~\cite{kollias2019face,kollias2021distribution} for simultaneous predicting facial Expressions (EXPR), Valence/Arousal (VA), and Action Units (AUs), along with 2) recognizing Compound Expressions (CE)~\cite{kollias2023multi} in real-world settings. The first tasks have been studied in ABAW-3~\cite{kollias2022abaw} and ABAW-4~\cite{kollias2023abaw}, while the latter problem has recently appeared in ABAW-6~\cite{kollias20246th}. The winners of these challenges achieve high accuracy by utilizing complex architectures and large ensembles~\cite{qiu2024learning,zhang2022multi}, which often require significant computational resources and extensive data processing capabilities. While these models excel in high-end hardware settings, their applicability in mobile or low-resource environments is limited~\cite{kharchevnikova2018neural,savchenko2022cvprw}. This restricts their deployment in many practical scenarios where real-time processing on personal devices is necessary, emphasizing the need for lightweight yet effective solutions.

Our approach to the ABAW-7 challenge focuses on addressing these issues by developing a streamlined pipeline based on efficient, frame-level facial feature extractors. These extractors, pre-trained in a multi-task setting to estimate valence-arousal and basic facial expressions, ensure high accuracy while maintaining low computational demands. To enhance privacy awareness, we utilize lightweight neural network architectures such as MT-EmotiDDAMFN, MT-EmotiMobileFaceNet~\cite{savchenko2024leveraging} and MT-EmotiEffNet~\cite{savchenko2023emotieffnets,savchenko2023hse}, which are designed to run efficiently on mobile devices. This eliminates the need to send facial video data to remote servers, safeguarding user privacy and ensuring compliance with stringent data protection regulations. Our methodology incorporates advanced post-processing techniques to refine the predictions of these models. By applying Gaussian or box filters to smooth the output scores and blending predictions from top-performing models, we significantly enhance facial expression recognition accuracy. Our experiments demonstrate that our final performance score in the multi-task learning challenge outperforms the baseline by a substantial margin, showcasing the efficacy of our approach in balancing accuracy, efficiency, and privacy.

The remaining part of this paper is structured as follows. Related works of MTL and CE competition participants are discussed in Section~\ref{sec:related}. Section~\ref{sec:methods} described the proposed approach. Thorough experimental results are discussed in Section~\ref{sec:exper}. Finally, Section~\ref{sec:concl} contains the conclusion and future works.

\section{Related Works}
\label{sec:related}
In the MTL task in the ABAW competition, it is required to assign an image of a facial frame to valence, arousal, one of 8 basic expressions,  and a subset of 12 AUs. The participants of ABAW-3 were not required to use the training set (s-Aff-Wild2), so most of them~\cite{Deng_2022_CVPR} were trained on larger AffWild2~\cite{kollias2019expression} set. Hence, their results on the validation set cannot be directly compared to participants of ABAW-7 and ABAW-4~\cite{kollias2023abaw} who were forced to refine their models on s-Aff-Wild2. Let us discuss only techniques proposed during the latter competition. 

The two-aspect information interaction model~\cite{sun2022two} represents interactions between sign vehicles and messages. The SS-MFAR~\cite{gera2022facial} extracts facial features using ResNet and leverages adaptive threshold for every class of facial expressions. The thresholds were estimated based on semi-supervised learning. The SMMEmotionNet was used to extract facial embeddings for an ensemble in the solution of the 6th place~\cite{mtl2022six}. At the same time, the hybrid CNN (Convolutional Neural Network)-Transformer~\cite{mtl2022fifth} with a fusion of ResNet-18 and a spatial transformer took the 5th place. Slightly better results have been obtained by the cross-attentive module and a facial graph that captures the association among action units~\cite{nguyen2022affective}. The EfficientNet model pre-trained in Multi-Task setting (MT-EmotiEffNet) took the third place~\cite{savchenko2023hse}. The top results have been achieved by the Masked Auto-Encoder (MAE) pretrained on unlabeled face images. For example, the EMMA ensemble of pre-trained MAE ViT (Vision Transformer) and CNN took the 2nd place~\cite{li2023affective}. The winner~\cite{zhang2022multi}  adopted ensembles of various temporal encoders, multi-task frameworks, and unsupervised (MAE-based) and supervised (IResNet/DenseNet-based) visual feature representation learning.


To encourage the study of cross-dataset settings, the ABAW-6 challenge~\cite{kollias20246th} presented the CE recognition problem from the C-EXPR database~\cite{kollias2023multi}, which does not contain the training set. Indeed, participants have to assign each frame of 56 videos into one of seven categories (Fearfully Surprised, Happily Surprised, Sadly Surprised, Disgustedly Surprised, Angrily Surprised, Sadly Fearful, and Sadly Angry).

The fifth place was achieved by the pre-trained visual language model that annotated a subset of the unlabeled data~\cite{wang2024zero}. Next, 5 CNNs have been fine-tuned using the generated labels. The audio-visual ensemble of three pre-trained models was utilized in~\cite{Ryumina_2024_CVPR}, where the predictions for basic expressions are weighted using the Dirichlet distribution and summarized to predict the compound expressions. Three different models (ResNet50, ViT and Multi-scale, and Local Attention Network) were fine-tuned on the dataset annotated with the same compound expressions~\cite{Yu_2024_CVPR}. Their combination with late fusion allowed the authors to reach third place. The lightweight MT-EmotiMobileFaceNet model with a simple sum of predictions corresponding to compound expressions reaches the second place~\cite{savchenko2024leveraging}. The winner~\cite{qiu2024learning} adopted the MAE pre-trained on a large facial dataset and the ViT encoder.  It was proposed to transform the task to a multi-label recognition task for basic emotions: the ViT encoder was finetuned on the part of the AffWild2 dataset to predict the probability of each basic emotion, which can be combined to make the final prediction for CE. 

Thus, most solutions of previous competitions adopted ensembles of deep models, which significantly limits their practical applications. Hence, in this paper, we try to simplify the decision-making pipeline for both tasks by leveraging only simple post-processing of lightweight models.

\section{Methods}\label{sec:methods}
The main task of this paper is to recognize the emotions of each video frame $X(t), t=t_1, t_2, ...,t_N$, where $1\le t_1< t_2< ...<t_N$ are the observed frame indices. We deal with both continuous and discrete models of emotions. In the former case, the most typical emotional space is the two-factor Russel's circumplex model of affect~\cite{russell1980circumplex} with VA-based emotional encoding. However, three- or even four-factor models have also been studied. Discrete representations have initially appeared as a set of basic expressions of Paul Ekman (anger, happiness, etc.). Moreover, detailed FACS (Facial Action Coding System)~\cite{ekman1978facial} with specific facial parts (AUs) is also widely used. Finally, due to the complexity of human emotions, it is also important to analyze compound expressions that have appeared simultaneously.

We consider two tasks of the ABAW-7 challenge. In the MTL competition, it is necessary to assign $X(t)$ to three emotional representations:
\begin{enumerate}
 \item Valence $V(t) \in [-1,1]$ and arousal $A(t) \in [-1,1]$ (multi-output regression task).
 \item Facial expression $c(t) \in \{1,..., C_{EXPR}\}$, where $C_{EXPR}=8$ is the total number of basic emotions: Neutral, Anger, Disgust, Fear, Happiness, Sadness, Surprise, and Other (multi-class classification).
 \item AUs $\mathbf{AU}(t)=[AU_1(t),...,AU_{C_{AU}}(t)]$, where $C_{AU} = 12$ is the total number of AUs and $AU_i(t) \in \{0,1\}$ (multi-label classification).
\end{enumerate}
 
\begin{figure}[t]
\centering
\includegraphics[width=\textwidth]{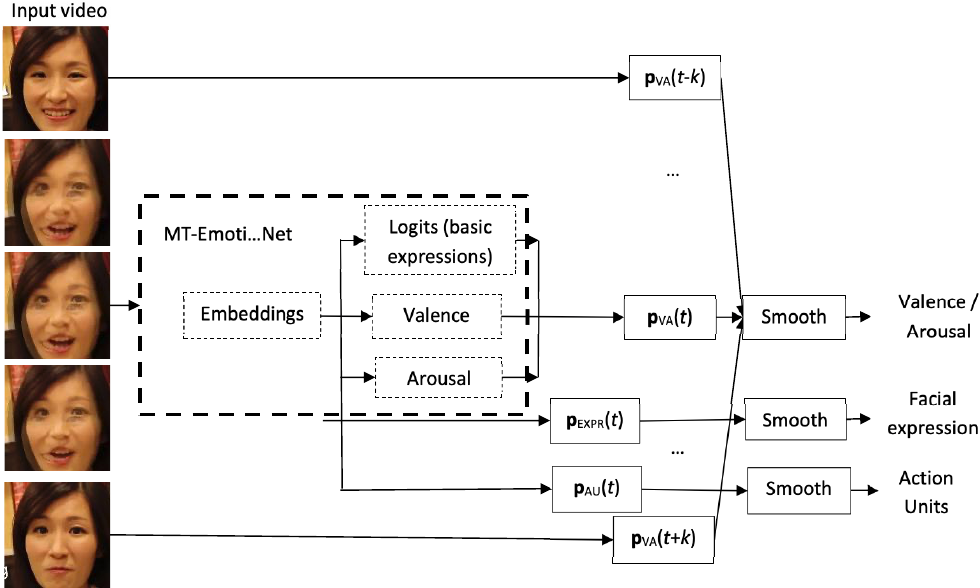}
\caption{Proposed pipeline}
\label{fig:1}
\end{figure}

This paper proposes a novel pipeline (Fig.~\ref{fig:1}) for the MTL challenge. The main part is the feature extractor backbones based on lightweight neural network architectures~\cite{savchenko2022cvprw}, such as EfficientNet-B0, MobileFaceNet, DDAMFN (Dual-Direction Attention Mixed Feature Network)~\cite{zhang2023dual} and MobileViT. We performed their training as described in~\cite{savchenko2024leveraging}, namely, firstly, pre-trained them to recognized faces from VGGFace2 dataset, and next fine-tune them on AffectNet to simultaneously classify static facial expressions and predict VA using multi-task loss~\cite{savchenko2023hse}, which is essentially a sum of weighted categorical cross-entropy for facial expressions and CCC (Concordance Correlation Coefficients) for valence and arousal. The resulting models (MT-EmotiEffNet, MT-EmotiDDAMFN, MT-EmotiMobileFaceNet, MT-EmotiMobileViT)~\cite{savchenko2024leveraging} are used to extract $D \gg 1$-dimensional facial embeddings $\mathbf{x}$ at the output of penultimate layer and 10-dimensional scores $\mathbf{s}$ (8 logits of emotions from AffectNet plus valence and arousal) at the output of the last layer.

Next, we train a simple feed-forward network with three output layers: 1) 2 hyperbolic tangent activations $\mathbf{p}_{VA} \in [-1,1]^2$; 2) softmax layer with $C_{EXPR}$ outputs ($\mathbf{p}_{EXPR} \in [0,1]^8$); and 3) $C_{AU}$ logistic sigmoids ($\mathbf{p}_{AU} \in \{0,1\}^{12}$)~\cite{savchenko2023hse}. The input of this model is a concatenation of $\mathbf{x}$ and $\mathbf{s}$. However, we empirically observed that VAs are predicted better using only logits, so we developed a special slice layer that extracts only 10 last inputs (logits $\mathbf{s}$) and feeds them into the VA dense layer. Moreover, we studied a straightforward blending of two top models indicated by upper indices $^{(1)}$ and $^{(2)}$:
\begin{align}\label{eq:blending}
\mathbf{p}^{(blend)}_{VA}(t)=w_{VA}\cdot\mathbf{p}^{(1)}_{VA} + (1-w_{VA})\cdot\mathbf{p}^{(2)}_{VA}, \\
\mathbf{p}^{(blend)}_{EXPR}(t)=w_{EXPR}\cdot\mathbf{p}^{(1)}_{EXPR} + (1-w_{EXPR})\cdot\mathbf{p}^{(2)}_{EXPR}, \\
\mathbf{p}^{(blend)}_{AU}(t)=w_{AU}\cdot\mathbf{p}^{(1)}_{AU} + (1-w_{AU})\cdot\mathbf{p}^{(2)}_{AU},
\end{align}
where the weights $w_{VA}$, $w_{EXPR}$ and $w_{AU}$ are chosen using available validation set.

As the values of sequential emotions should be smooth, we propose to perform post-processing of our predictions using simple filters. In general, dynamic changes of facial expressions, VA and AUs may be significantly different, so we use separate kernel sizes $k_{EXPR}$, $k_{VA}$, and $k_{AU}$. In practice, we observed that smoothing decreases the quality of AU detection, so we set $k_{AU}=0$. Nevertheless, we obtain a sequence of predictions for each task and compute either box filter (simple average):
\begin{equation}\label{eq:box}
\overline{\mathbf{p}}^{(avg)}(t)=\sum \limits_{|t_i-t| \le k}{\mathbf{p}(t_i)},
\end{equation}
or Gaussian filter with the smoothing factor (variance) $\sigma^2$:
\begin{equation}\label{eq:gauss}
\overline{\mathbf{p}}^{(Gauss)}(t)=\frac{\sum \limits_{|t_i-t| \le k}{(\frac{\exp(-(t-t_i)^2)}{2\sigma^2}\mathbf{p}(t_i))}}{\sum \limits_{t_i \in \Delta T(t)}{\frac{\exp(-(t-t_i)^2)}{2\sigma^2}}}.
\end{equation}

The final decision for the VA prediction task is $\overline{\mathbf{p}}_{VA}$. The index that corresponds to the maximal score $\overline{\mathbf{p}}_{EXPR}$ is returned for EXPR classification. The final AU scores are obtained by a hard decision rule in which scores $\overline{\mathbf{p}}_{AU}$ are compared with fixed thresholds $t_{AU}$. We consider two types of thresholds: fixed threshold $t_{AU}=0.5$ and the best thresholds $t^*_{AU}$, achieving the top macro-averaged F1-score on the validation set.

\begin{table}[tb]
\caption{Results of various feature extractors for the MTL challenge. The best result is marked in bold.}
\label{table:mtl_ablation}
  \centering
  \begin{tabular}{@{}llllll@{}}
\toprule
 Model &  Dataset& $P_{VA}$ & $P_{EXPR}$ & $P_{AU}$ & $P_{MTL}$  \\
\midrule
MT-EmotiMobileFaceNet& cropped & 0.4267 & 0.2784 & 0.4731 & 1.1782\\
 & cropped\_aligned & 0.4503 & 0.2870& 0.4891 & 1.2264\\
MT-MobileViT & cropped & 0.4123 & 0.3051 & 0.4689 & 1.1863\\ 
& cropped\_aligned & 0.4379 & 0.3146 & 0.4817 & 1.2342 \\
DDAMFN & cropped & 0.4505 & 0.3045 & 0.4773 & 1.2324\\
& cropped\_aligned & 0.4691 & 0.3148 & 0.4944& 1.2783 \\
MT-EmotiEffNet-B0 & cropped & 0.4469 & 0.3308 & 0.4943 & 1.2719\\
& cropped\_aligned & 0.4433 & \bf 0.3422 & \bf 0.5040 & 1.2896 \\
MT-EmotiDDAMFN & cropped & 0.4479 & 0.2864 & 0.4894 & 1.2238\\
& cropped\_aligned &\bf 0.4826 & 0.3396 & 0.4906& \bf 1.3128\\
\bottomrule
\end{tabular}
\end{table}

\begin{table}[tb]
\caption{Results of pre-trained models on the s-Aff-Wild2’s validation set. }
 \label{tab:expr_pretrained}
 \centering
 \begin{tabular}{@{}l|llll|lll@{}}
 \toprule
 & \multicolumn{4}{c|}{EXPR} & \multicolumn{3}{c}{VA} \\
& \multicolumn{2}{c}{all classes}& \multicolumn{2}{c|}{w/o ``Other''} \\
Model & F1-score & Accuracy & F1-score & Accuracy & CCC\_V & CCC\_A& $P_{VA}$ \\
 \midrule
MT-EmotiEffNet-B0 & 0.2836 & 0.2805 & 0.3880 & 0.4006 & 0.4251 & 0.2444 & 0.3348\\
MT-EmotiDDAMFN & 0.2932 & 0.3097 & 0.3934 & 0.4364 &0.3816 & 0.2277 & 0.3047\\
 \bottomrule
 \end{tabular}
\end{table}

We modified the pipeline originally described in~\cite{savchenko2024leveraging} for the CE classification task. In particular, we perform face detection in each frame using the RetinaFace~\cite{deng2020retinaface} and Mediapipe~\cite{lugaresi2019mediapipe}. Next, we feed each facial image into an emotional neural network and compute probability scores $\mathbf{p}$ for 8 basic emotions from AffectNet~\cite{mollahosseini2017affectnet}. We study two possibilities, namely, leveraging lightweight models pre-trained on AffectNet~\cite{savchenko2024leveraging} and using them to extract features for a feed-forward neural network with one hidden layer trained on the frame-wise expression classification task from ABAW-6~\cite{kollias20246th}. In the latter case, a straightforward multi-class classification problem was solved with softmax activation and weighted categorical cross-entropy loss using either the official training part or the train and validation set concatenation. Moreover, we implemented the idea of the winner of the previous edition of this challenge~\cite{qiu2024learning}. We trained a model for multi-label classification with sigmoid activations and weighed binary cross-entropy loss.

Next, the scores $\mathbf{p}$ are aggregated for two classes from the compound expressions. We use three different average functions: arithmetic mean (A-mean), geometric mean (G-mean), and harmonic mean (H-mean). As one frame may contain several detected faces, we examine two options: 1) compute the simple average of CE predictions inside one frame, or 2) choose only the largest face and select its predictions. Finally, we perform smoothing of sequential predictions with kernel size $k$ using either box (\ref{eq:box}) or Gaussian (\ref{eq:gauss}) filters and return the label corresponding to the maximal smoothed score. 

\section{Experiments}\label{sec:exper}
\subsection{Multi-Task Learning Challenge}

\begin{figure}[tb]
  \centering
  \begin{subfigure}{0.47\textwidth}
    \includegraphics[width=\textwidth]{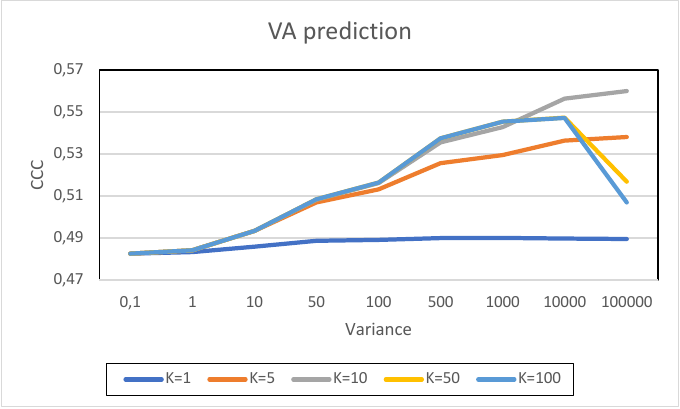}
    \caption{}
    \label{fig:va_mtddamfn}
\end{subfigure}
\hfill
\begin{subfigure}{0.47\textwidth}
    \includegraphics[width=\textwidth]{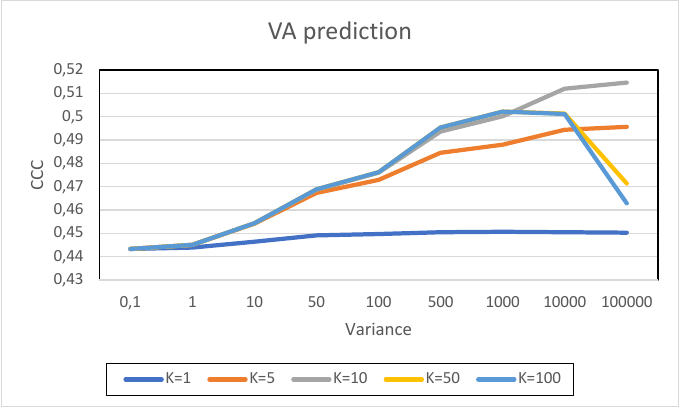}
    \caption{}
    \label{fig:va_mtenetb0}
\end{subfigure}
   \caption{Dependence of the average CCC of VA prediction on smoothing variance $\sigma^2$: (a) MT-EmotiDDAMFN, (b) MT-EmotiEffNet-B0.}
   \label{fig:mtl_va_smooth}
\end{figure}

\begin{figure}[tb]
  \centering
  \begin{subfigure}{0.47\textwidth}
    \includegraphics[width=\textwidth]{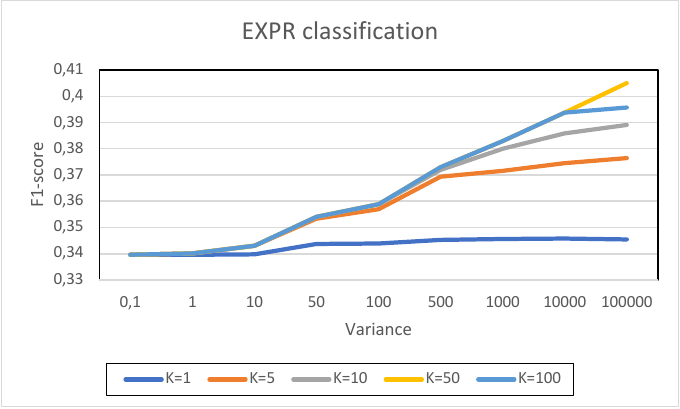}
    \caption{}
    \label{fig:expr_mtddamfn}
\end{subfigure}
\hfill
\begin{subfigure}{0.47\textwidth}
    \includegraphics[width=\textwidth]{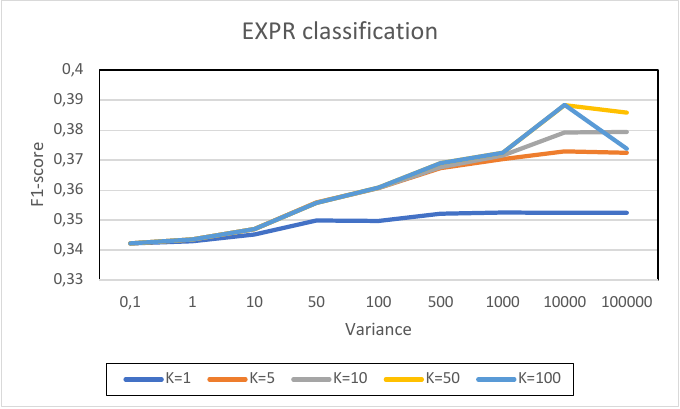}
    \caption{}
    \label{fig:expr_mtenetb0}
\end{subfigure}
   \caption{Dependence of the F1-score of EXPR classification on smoothing variance $\sigma^2$: (a) MT-EmotiDDAMFN, (b) MT-EmotiEffNet-B0.}
   \label{fig:mtl_expr_smooth}
\end{figure}

\begin{figure}[tb]
  \centering
  \begin{subfigure}{0.47\textwidth}
    \includegraphics[width=\textwidth]{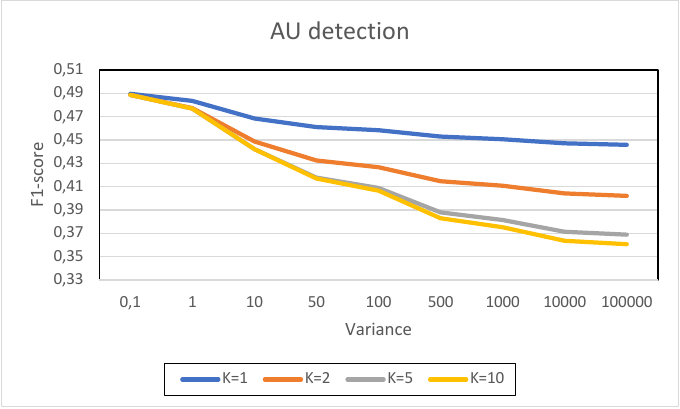}
    \caption{}
    \label{fig:au_mtddamfn}
\end{subfigure}
\hfill
\begin{subfigure}{0.47\textwidth}
    \includegraphics[width=\textwidth]{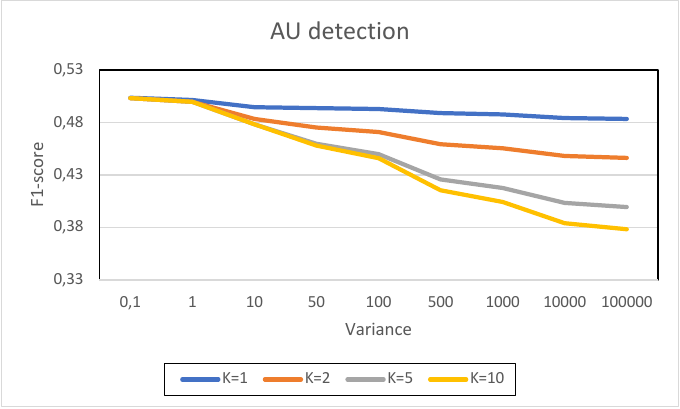}
    \caption{}
    \label{fig:au_mtenetb0}
\end{subfigure}
   \caption{Dependence of the F1-score of AU detection on smoothing variance $\sigma^2$: (a) MT-EmotiDDAMFN, (b) MT-EmotiEffNet-B0.}
   \label{fig:mtl_au_smooth}
\end{figure}

\begin{figure}[tb]
  \centering
  \begin{subfigure}{0.47\textwidth}
    \includegraphics[width=\textwidth]{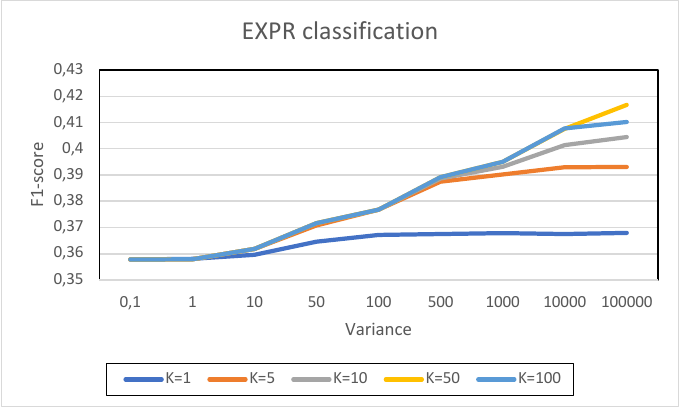}
    \caption{}
    \label{fig:expr_blending}
\end{subfigure}
\hfill
\begin{subfigure}{0.47\textwidth}
    \includegraphics[width=\textwidth]{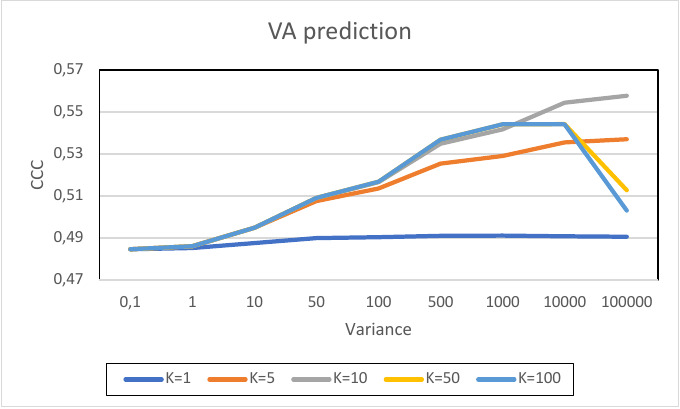}
    \caption{}
    \label{fig:va_blending}
\end{subfigure}
   \caption{Dependence of blending results on smoothing variance $\sigma^2$: (a) EXPR classification, (b) VA prediction.}
   \label{fig:mtl_blending_smooth}
\end{figure}

\begin{table}[tb]
\caption{Class-level F1 score for Valence-Arousal prediction}
\label{tab:best_va}
 \centering
 \begin{tabular}{@{}ll|ll@{}} 
 \toprule
Model& Details & Valence & Arousal\\
 \midrule
MT- & frame-level & 0.4433 & 0.3422 \\
EmotiEffNet & smoothing & 0.5268 & 0.5025\\
\hline
MT-& frame-level& 0.4926 & 0.4725\\
DDAMFN& smoothing& 0.5603 & 0.5597 \\
\hline
MT-EmotiEffNet& frame-level& 0.4968 & 0.4726\\
+MT-EmotiDDAMFN& smoothing& 0.5598 & 0.5558\\
\bottomrule
 \end{tabular}
\end{table}

\begin{table}[tb]
\caption{Class-level F1 score for the FER task}
\label{tab:best_expr}
 \centering
 \resizebox{\textwidth}{!}{ 
 \begin{tabular}{@{}ll|llllllll@{}} 
 \toprule
Model& Details & \multicolumn{8}{c}{Expression} \\
& & Neutral & Anger & Disgust & Fear & Happy & Sad & Surprise & Other \\
 \midrule
MT- & frame-level & 0.3373 & 0.2111 & 0.4404 & 0.2279 & 0.5013 & 0.4801 & 0.2396 & 0.3000\\
EmotiEffNet & smoothing & 0.3789 & 0.2051 & 0.5058 & 0.2924 & 0.5450 & 0.5956 & 0.2357 & 0.3487\\
\hline
MT-& frame-level& 0.3949 & 0.2101 & 0.3011 & 0.2597 & 0.4796 & 0.5043 & 0.2639 & 0.3035\\
DDAMFN& smoothing& 0.4484 & 0.2117 & 0.4165 & 0.3056 & 0.5014 & 0.7631 & 0.2386 & 0.3551\\
\hline
MT-EmotiEffNet& frame-level& 0.3944 & 0.2152 & 0.3992 & 0.2621 & 0.4988 & 0.5157 & 0.2677 & 0.3095\\
+MT-EmotiDDAMFN& smoothing& 0.4338 & 0.2268 & 0.5173 & 0.3025 & 0.5131 & 0.7332 & 0.2441 & 0.3632\\
\bottomrule
 \end{tabular}
 }
\end{table}

\begin{table}[tb]
\caption{Detailed results for Action Unit detection}
\label{tab:best_au}
 \centering
 \resizebox{\textwidth}{!}{ 
 \begin{tabular}{@{}ll|llllllllllll@{}} 
 \toprule
Model& & \multicolumn{12}{c}{Action Unit \#} \\
& & 1  &  2 &  4 &  6 &  7 &  10 &  12 &  15 &  23 &  24 &  25 &  26 \\
 \midrule
MT-& $P_{AU}(t_{AU}=0.5)$ & 0.59 & 0.44 & 0.56 & 0.60 & 0.73 & 0.71 & 0.66 & 0.23 & 0.17 & 0.16 & 0.85 & 0.36 \\
\cline{2-14}
EmotiEffNet& Best $t^*_{AU}$ & 0.6 & 0.7 & 0.6 & 0.4 & 0.4 & 0.5 & 0.5 & 0.7 & 0.8 & 0.8 & 0.1 & 0.6 \\
& $P_{AU}(t^*_{AU})$ & 0.61 & 0.44 & 0.60 & 0.59 & 0.72 & 0.71 & 0.68 & 0.23 & 0.16 & 0.20 & 0.88 & 0.36 \\
\hline 
MT-& $P_{AU}(t_{AU}=0.5)$ & 0.59 & 0.38 & 0.59 & 0.58 & 0.72 & 0.71 & 0.68 & 0.19 & 0.11 & 0.15 & 0.82 & 0.35 \\
\cline{2-14}
DDAMFN& Best $t^*_{AU}$ & 0.6 & 0.7 & 0.6 & 0.4 & 0.4 & 0.5 & 0.5 & 0.7 & 0.8 & 0.8 & 0.1 & 0.6 \\
& $P_{AU}(t^*_{AU})$ & 0.61 & 0.44 & 0.60 & 0.59 & 0.72 & 0.71 & 0.68 & 0.23 & 0.16 & 0.20 & 0.88 & 0.36 \\
\hline 
MT-& $P_{AU}(t_{AU}=0.5)$ & 0.61 & 0.44 & 0.60 & 0.59 & 0.72 & 0.71 & 0.68 & 0.23 & 0.16 & 0.20& 0.88 & 0.36 \\
\cline{2-14}
EmotiEffNet+& Best $t^*_{AU}$ & 0.6 & 0.7 & 0.6 & 0.4 & 0.4 & 0.5 & 0.5 & 0.7 & 0.8 & 0.8 & 0.1 & 0.6 \\
MT-EmotiDDAMFN& $P_{AU}(t^*_{AU})$ & 0.62 & 0.44 & 0.59 & 0.61 & 0.74 & 0.72 & 0.69 & 0.26 & 0.17 & 0.19& 0.85 & 0.37 \\ 
\bottomrule
 \end{tabular}
 }
\end{table}

In this subsection, we provide an ablation study of our approach for the MTL challenge using official training and validation sets from s-Aff-Wild2~\cite{kollias2022abaw}. Some labels are missed, so 142,333 training facial frames have only 103,917 values of Valence / Arousal, 90,645 expression labels, and 103,316 AUs. The validation set consists of 26,876 faces: all AU and VA are available, but only 15,440 expressions are known. We use official performance measure ($P_{MTL}=P_{VA}+P_{EXPR}+P_{AU}$), which is the sum of the mean CCC of valence and arousal ($P_{VA}=(CCC_V+CCC_A)/2$); the macro-averaged F1-score across all 8 expression categories ($P_{EXPR}$); the average F1-score across all 12 AUs ($P_{AU}$).

The training results on the cropped and cropped\_aligned official sets are shown in Table~\ref{table:mtl_ablation}. As one can notice, the latter set is much better for our models. Moreover, the best results are obtained by MT-EmotiEffNet-B0~\cite{savchenko2023hse} and MT-EmotiDDAMFN~\cite{savchenko2024leveraging}, which will be considered in the remaining experiments. 

The results of their pre-trained versions without refinement on s-Aff-Wild2 are presented in Table~\ref{tab:expr_pretrained}. As there is no Other category in AffectNet and Contempt emotion in s-Aff-Wild2, we implemented two techniques for matching expressions: assigning all ``Contempt'' predictions as ``Other'' or excluding the ``Contempt'' class from predictions and ``Other'' from the validation dataset. Though the performance measures on EXPR classification and VA prediction tasks are slightly worse when compared to training a feed-forward net on the s-Aff-Wild2 (Table~\ref{table:mtl_ablation}), they still demonstrate promising results, especially if rather broad and unspecific ``Other'' category is ignored. It is also remarkable that though the pre-trained MT-EmotiEffNet-B0 achieves higher mean CCC $P_{VA}$ when compared to MT-EmotiDDAMFN, the latter can be trained to achieve much better final $P_{VA}$ (0.4826 vs 0.4433, see Table~\ref{table:mtl_ablation}).

In the next experiments, we study the Gaussian smoothing (\ref{eq:gauss}) of predictions (Fig.~\ref{fig:mtl_va_smooth},~\ref{fig:mtl_expr_smooth} and~\ref{fig:mtl_au_smooth}). As one can see, smoothing does not work for AU detection but can significantly improve the results for other tasks: up to 0.06 difference in F1-score for EXPR classification and up to 0.06 difference in mean CCC for VA prediction. Moreover, the smoothing works nicely even for blending the best models (Fig.~\ref{fig:mtl_blending_smooth}). The detailed results of our approach for each category of VA, EXPR, and AU are presented in Table~\ref{tab:best_va}, Table~\ref{tab:best_expr} and Table~\ref{tab:best_au}, respectively. 

\begin{table}[tb]
\caption{Validation results for the MTL challenge}
\label{table:mtl_validation}
  \centering
  \begin{tabular}{@{}llllll@{}} 
\toprule
 Model & Details & $P_{VA}$ & $P_{EXPR}$ & $P_{AU}$ & $P_{MTL}$  \\
\midrule
\multicolumn{2}{c}{Baseline VGGFace~\cite{kollias2023abaw}}& -& - & - & 0.32\\
\multicolumn{2}{c}{Hybrid CNN-Transformer~\cite{mtl2022fifth}}& - & - & - & 0.981\\
\multicolumn{2}{c}{SMMEmotionNet~\cite{nguyen2022affective}} & 0.3648 & 0.2617 & 0.4737 & 1.1002\\
\multicolumn{2}{c}{Two-Aspect Information Interaction~\cite{sun2022two}}& 0.515 & 0.207 & 0.385 & 1.107\\
\multicolumn{2}{c}{SS-MFAR~\cite{gera2022facial}}& 0.397 & 0.235 & 0.493 & 1.125\\
\multicolumn{2}{c}{EfficientNet-B2~\cite{savchenko2022cvprw}}& 0.384 & 0.302 & 0.461& 1.147 \\
\multicolumn{2}{c}{MAE+ViT~\cite{li2023affective}} & 0.4588 & 0.3028 & 0.5054 & 1.2671\\
\multicolumn{2}{c}{Cross-attentive module~\cite{nguyen2022affective}} & 0.499 & 0.333 & 0.456 & 1.288\\
\multicolumn{2}{c}{MT-EmotiEffNet + OpenFace~\cite{savchenko2023hse}} & 0.447 & 0.357 & 0.496& 1.300 \\
\multicolumn{2}{c}{MAE+Transformer~\cite{zhang2022multi}} & 0.6547 & 0.3901 & 0.5579 & 1.6027\\
\hline
& frame-level, $t_{AU}=0.5$ & 0.4433 & 0.3422 & 0.5040 & 1.2896 \\
MT-EmotiEffNet-B0 & smoothing, $t_{AU}=0.5$ & 0.5147 & 0.3884 & 0.5040 & 1.4071 \\
& smoothing, best $t^*_{AU}$ & 0.5147 & 0.3884 & 0.5170 & 1.4201 \\ \hline
& frame-level, $t_{AU}=0.5$ & 0.4826 & 0.3396 & 0.4906& 1.3128 \\
MT-EmotiDDAMFN & smoothing, $t_{AU}=0.5$ & 0.5600 & 0.4051 & 0.4906 & 1.4557 \\
& smoothing, best $t^*_{AU}$ & 0.5600 & 0.4051 & 0.5154 & 1.4805 \\ \hline
MT-EmotiEffNet-B0  & frame-level & 0.4847 & 0.3578 & 0.5194 & 1.3619 \\
+ MT-EmotiDDAMFN & smoothing, $t_{AU}=0.5$ & 0.5578 & 0.4168 & 0.5194 & 1.4939 \\
\bottomrule
\end{tabular}
\end{table}

The best results of our approach compared to known results for the validation set of the MTL challenge are presented in Table~\ref{table:mtl_validation}. We significantly improve the previous best metric $P_{MTL}$ of MT-EmotiEffNet~\cite{savchenko2023hse}: from 1.30 to 1.42. Moreover, the usage of recently introduced MT-EmotiDDAMFN in a simple blending ensemble (\ref{eq:blending}) wit Gaussian smoothing (\ref{eq:gauss}) makes it possible to achieve $P_{MTL}=1.49$, which is 5-times better than the baseline VGGFace~\cite{kollias2023abaw} (0.32) and only slightly worse than the first place of previous challenge~\cite{zhang2022multi}.

\subsection{Compound Expression Recognition Challenge}

\begin{table}[tb]
  \caption{Results of pre-training on AffWild2 for EXPR classification}
  \label{tab:expr_pretrain_affwild}
  \centering
  \begin{tabular}{@{}lllll@{}}
    \toprule
    & \multicolumn{2}{c}{Multi-class} &\multicolumn{2}{c}{Multi-label} \\
	Model & Accuracy & F1-score & Accuracy & F1-score\\
    \midrule
MT-EmotiEffNet-B0 & 0.4874 & 0.4111 & 0.4906 & 0.3890 \\
EmotiEffNet-B0 & 0.5852 & 0.4560 & 0.5894 & 0.4697 \\
MT-EmotiMobileFaceNet & 0.5362 & 0.3968& 0.5546 & 0.4250 \\
MT-EmotiDDAMFN & 0.5550 & 0.3933& 0.5486 & 0.4211 \\
DDAMFNet & 0.5350 & 0.3962 & 0.5081 & 0.3587 \\
MT-MobileViT & 0.5735 & 0.4347 & 0.5716 & 0.4210 \\
  \bottomrule
  \end{tabular}
\end{table}

\begin{table}[tb]
  \caption{KL divergence for CE recognition. Here, A-mean, G-mean, and H-mean stand for Arithmetic mean, Geometric mean, and Harmonic mean}
  \label{tab:ce}
  \centering
  \resizebox{\textwidth}{!}{ 
  \begin{tabular}{@{}lllllll@{}}
    \toprule
    & \multicolumn{3}{c}{All faces} &\multicolumn{3}{c}{Largest face} \\
	Model & A-mean & G-mean & H-mean &	A-mean & G-mean & H-mean \\
    \midrule
MT-EmotiMobileFaceNet & 0.060 & 0.102 & 0.123 & 0.052 & 0.106 & 0.144 \\
EmotiEffNet-B0 & 0.119 & 0.209 & 0.269 & 0.123 & 0.203 & 0.261 \\
MT-EmotiEffNet-B0 (EXPR-ft), train only & 0.193 & 0.192 & 0.189 & 0.173 & 0.174 & 0.175 \\
EmotiEffNet-B0 (EXPR-ft), multilabel & 0.545 & 0.541 & 0.511 & 0.504 & 0.505 & 0.504 \\
EmotiEffNet-B0 (EXPR-ft), train only & 0.422 & 0.439 & 0.416 & 0.396 & 0.400 & 0.404 \\
EmotiEffNet-B0 (EXPR-ft), train+val & 0.236 & 0.241 & 0.209 & 0.210 & 0.209 & 0.211\\
  \bottomrule
  \end{tabular}
  }
\end{table}

\begin{figure}[tb]
  \centering
  \begin{subfigure}{0.52\textwidth}
    \includegraphics[width=\textwidth]{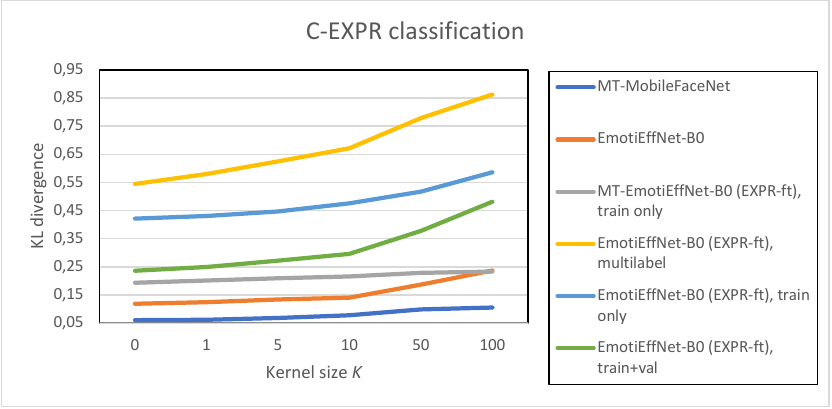}
    \caption{}
    \label{fig:cexpr_models}
\end{subfigure}
\hfill
\begin{subfigure}{0.42\textwidth}
    \includegraphics[width=\textwidth]{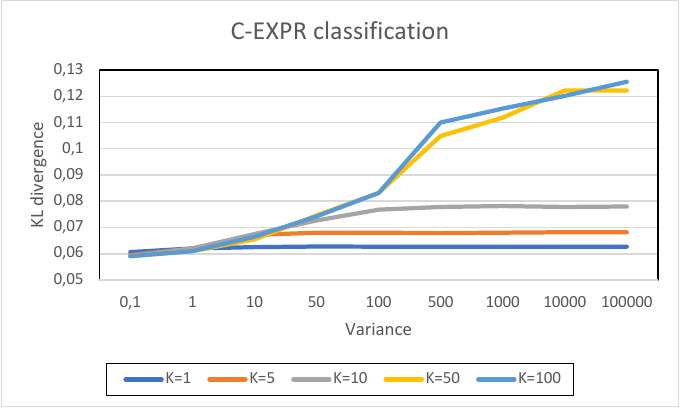}
    \caption{}
    \label{fig:cexpr_smooth}
\end{subfigure}
   \caption{The Kullback-Leibler divergence between real and predicted class probabilities for CE recognition: (a) various feature extractors; (b) Dependence on smoothing variance $\sigma^2$, MT-EmotiMobileFaceNet feature extractor.}
   \label{fig:cexpr}
\end{figure}

\begin{figure}[tb]
  \centering
  \includegraphics[width=0.75\linewidth]{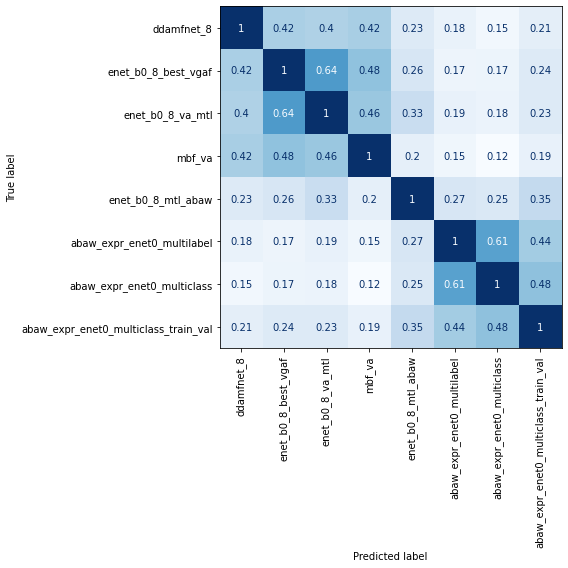}
   \caption{Kappa Cohen scores for CE predictions}
   \label{fig:ce_kappa}
\end{figure}

In this Subsection, we present the results of the CE classification challenge. At first, we present the results of refinement of EXPR classification feed-forward networks using AffWild2 (Table~\ref{tab:expr_pretrain_affwild}). The best F1-score is obtained by EmotiEffNet-B0, while the multi-label classification problem lets us obtain a more accurate network. In the remaining part of this section, we will denote these models using the suffix ``(EXPR\_ft)''.

Speaking about results on CE recognition, there are no training/validation sets for this task. Hence, we borrow the approach from~\cite{savchenko2024leveraging} and estimate the resulting class balance by computing the Kullback-Leibler (KL) divergence between actual and predicted class probabilities (Table~\ref{tab:ce}, Fig.~\ref{fig:cexpr_models}). The actual frequencies of each category were taken from original paper~\cite{kollias2023multi}: Fearfully Surprised (14445 frames), Happily Surprised (24915), Sadly Surprised (10780), Disgustedly Surprised (10637), Angrily Surprised (10535), Sadly Fearful (10112), and Sadly Angry (8878). Here, the MT-EmotiMobileFaceNet for the largest face in a frame and arithmetic mean of compound class probabilities seems to be the best candidate for submission. However, the Gaussian smoothing of its predictions slightly increases the KL divergence (Fig.~\ref{fig:cexpr_smooth}). Cohen's kappa coefficient estimates the inter-rater reliability (Fig.~\ref{fig:ce_kappa}). This figure has two parts: the top-left corner of models pre-trained on AffectNet and the left-bottom corner for EXPR classifiers trained on AffWild2.

\begin{table}[tb]
  \caption{F1-score of CE recognition on the test set}
  \label{tab:ce_test}
  \centering
  \begin{tabular}{@{}ll@{}}
    \toprule
Model & F1-score\\
    \midrule
5 CNNs and visual language model~\cite{wang2024zero} & 0.1845 \\
Audio-visual ensemble~\cite{Ryumina_2024_CVPR} & 0.2201\\
ResNet50, ViT, Multi-scale and Local Attention Network~\cite{Yu_2024_CVPR} & 0.2240\\
MAE+ViT~\cite{qiu2024learning} & 0.5526\\
\hline
MT-EmotiEffNet~\cite{savchenko2024leveraging} & 0.2341\\
MT-EmotiEffNet (EXPR\_ft)~\cite{savchenko2024leveraging} & 0.2580\\
EmotiEffNet-B0~\cite{savchenko2024leveraging} & 0.2625\\
MT-EmotiMobileFaceNet~\cite{savchenko2024leveraging} & 0.2708 \\
\hline
Ours (MT-EmotiEffNet, box filter) & 0.2602\\
Ours (MT-EmotiEffNet, EXPR\_ft, box filter) & 0.3067 \\
Ours (EmotiEffNet-B0, box filter) & 0.2890\\
Ours (MT-EmotiMobileFaceNet, box filter) & 0.3146\\
  \bottomrule
  \end{tabular}
\end{table}

The results on the test set (Table~\ref{tab:ce_test}) demonstrate the superiority of using our post-processing. Indeed, we increase the F1-score to 4.5\% (from 0.2708 to 0.3146) for the best MT-EmotiMobileFaceNet model.

\section{Conclusion}\label{sec:concl}
In this paper, we proposed a novel pipeline (Fig.~\ref{fig:1}) for efficient solutions to various emotion recognition problems. We use lightweight neural network feature extractors pre-trained on AffectNet to simultaneously recognize basic expressions and predict valence and arousal.  Our method does not need to fine-tune the neural network model on a new dataset. Experiments on datasets from ABAW-6 competition show that even pre-trained models demonstrate reasonable performance (Table~\ref{tab:expr_pretrained}), and training a simple feed-forward neural network with 1 hidden layer makes it possible to reach even better results (Table~\ref{table:mtl_ablation}). The post-processing of frame-wise predictions with Gaussian (\ref{eq:gauss}) or box filters (\ref{eq:box}) and blending of two best models (\ref{eq:blending}) improves the F1-score of facial expression classification up to 7\% (Fig.~\ref{fig:mtl_expr_smooth}), while the mean CCC of VA prediction is up to 1.25-times better (Fig.~\ref{fig:mtl_va_smooth}) than the frame-level predictions for both MTL (Table~\ref{table:mtl_validation}) and CE challenges (Table~\ref{tab:ce_test}). As a result, our final performance score on the validation set from the MTL task is 4.5 times higher than the baseline (1.494 vs 0.32). 

The source code to reproduce our experiments is available at\footnote{\url{https://github.com/HSE-asavchenko/face-emotion-recognition/tree/main/src/ABAW/ABAW7}}. In the future, it is necessary to continue the study of pre-trained models for feature extractors similarly to the winners of previous challenge who successfully trained an effective facial feature extractor using MAE and ViT~\cite{qiu2024learning,zhang2022multi}. Moreover, it is possible to integrate the pre-trained speaker emotion recognition models~\cite{Ryumina_2024_CVPR} to improve the accuracy of our pipeline.

\subsubsection{Acknowledgements} The article was prepared within the framework of the Basic Research Program at the National Research University Higher School of Economics (HSE).

%
%
\bibliographystyle{splncs04}
\bibliography{main}
\end{document}